RESEARCH ARTICLE

# ArabFake: A Multitask Deep Learning Framework for Arabic Fake News Detection, Categorization, and Risk Prediction

**AHMED MAHER KHAFAGA SHEHATA[1], MOHAMMED NASSER AL-SUQRI[1], NOUR ELDIN MOHAMED ELSHAIEKH OSMAN[1], FATEN HAMAD[1,2], YOUSUF NASSER ALHUSAINI[3], (Member, IEEE), AND AHMED MAHFOUZ[3,4]**

[1]Information Studies Department, Sultan Qaboos University, Seeb 123, Oman
[2]Library and information Science Department, University of Jordan, Amman, Jordan
[3]Faculty of Computer Studies, Arab Open University, Muscat 121, Oman
[4]Computer Science Department, Minia University, Minya 1596, Egypt

Corresponding author: Ahmed Maher Khafaga Shehata (a.shehata@squ.edu.om)

This work was supported by the Ministry of Higher Education, Research and Innovation (MoHERI) of the Sultanate of Oman under the Block Funding Program through MoHERI Block Funding under Agreement MoHERI/BFP/SQU/01/2021.

**ABSTRACT** The spread of fake news among Arabic media including social media represents a great challenge to the integrity of information and the trust of the public in it. In this paper, we introduce a comprehensive deep-learning framework, named ArabFake, that goes beyond the binary classification on Arabic fake news detection. ArabFake, built over MARBERTv2 (a state-of-the-art model for multi-dialectal Arabic tweets), proficiently address the complexity of the Arabic language while performing three unified tasks which are fake news detection, content categorization and its risk assessment. The framework promotes efficiency and performance both by enabling multi-task learning through shared knowledge representation across tasks. In order to facilitate development and evaluation, we present the ArabFake Dataset consisting of 2,495 manually labelled news items with labels that are verified by experts regarding fake news categories and risk levels. ArabFake demonstrates robust performance, achieving an F1 score of 94.12% for fake news detection, 84.92% for categorization, and 88.91% for risk zone assessment, highlighting its reliability and effectiveness across multiple tasks. We improve interpretability and extract insight into manipulative techniques by integrating valence scoring as part of the framework that emphasizes misleading linguistic cues used to disseminate fake news within the produced image. The results show that ArabFake is a holistic Arabic fake news detection framework that has practical implications on news organizations and fact checking projects.

## I. INTRODUCTION

In today's digital age, fake news spreads quickly online, causing doubts about reliable information and trust. This problem is even more serious in Arabic, where the language's unique features make it harder to spot fake news. Detecting and confronting fake news in Arabic is considered a very complex matter due to the seriousness of the news and its harmful effects. Therefore, governments are making great efforts to address misinformation and respond to it to prevent the resulting negative effects, such as the exacerbation of the problem due to the spread and the amplification of misinformation on social media networks.

Previous studies show that misinformation is diverse in type and nature, making identifying and discovering it extremely difficult. Also, the flood of misinformation has exhausted countries and exposed people worldwide to dangerous and harmful information. Fake news and misinformation spread across social media networks in all languages, not limiting the problem to one language or one country but extending its impact to people worldwide. Arabic speakers are no exception to fake news, as social networks are flooded with misleading or inaccurate information related to COVID-19 [1], [2].

The propagation of misinformation on the Internet and social media has shown a severe adverse effect, including confusing and manipulating public opinions, hindering the government's efforts to respond to the pandemic, changing people's behavior, and, in some cases, posing many threats to society [3]. Due to the COVID-19 pandemic, governments across the globe have had to exert additional efforts to prevent the spread of fake news and manage the pandemic itself. These efforts include publishing facts through social media, media campaigns, publishing fact-checking tools, implementing new legislation, and enhancing citizens' information literacy skills [4], [5].

Several studies have explored the problem of misinformation before and during COVID-19, confirming that people generally fail to recognize misinformation because they lack appropriate health information literacy skills and other factors such as education, trust, and personal beliefs [6], [7]. Additionally, the COVID-19 outbreak has revealed that the problem of misinformation is challenging for many reasons. Social, emotional, and cultural factors affect the absorption of misinformation, hindering efforts to stop the negative impact of this type of information [8].

Detecting misinformation on social networks has been difficult for both Internet users in general and for researchers in the field of natural language processing in particular for several years. The problem can be summarized in that misleading information varies in terms of message, purpose, nature, and language, which makes discovering this type of information difficult as it is a mixture of fabricated and correct information. As a result, computer science professionals have taken several measures to formulate strategies for detecting fake news. As a result of these efforts, many algorithms and models have been developed that take advantage of artificial intelligence to distinguish, identify, and extract fake or misleading news from social media networks and the Internet [9], [10], [11], [12], [13]. It is worth noting that studies have revealed that the Arabic language has received less attention in this field, and this is due to the characteristics that are unique to the Arabic language in terms of complexity, which require the presence of more complex algorithms that can deal with it [14], [15].

While prior works have addressed the problem of Fake News Detection in Arabic, existing methods rely essentially on binary classification and do not fully account for specificities in similarities, unlike those evolved by reading left to right languages [12], [16]. The existing models also suffer from the issue of interpretability and do not use multi-task learning for proper detection, categorization & assessment of the risk of fake news. In addition, Arabic fake news lacks large-scale, high-quality labeled datasets, which impedes the design and evaluation of more advanced models. To overcome these limitations, we present ArabFake: a unified language-aware and interpretable representation learning method designed for Arabic fake news detection through multitask reasoning. Second, we create a novel dataset with expert annotations. The release of the ArabFake Dataset is therefore a crucial resource for researchers interested in this field. Our study has evaluated the performance of ArabFake on a variety of datasets and tasks that shows state-of-the-art results when benchmarked with other models, which can greatly contribute towards fake news detection in Arabic-related research.

The ArabFake model is built atop multi-task learning (MTL), enabling it to simultaneously foreground fake news, gauge the relative strength of such text as well as label them [17]. In addition, the ArabFake framework is computationally efficient and scalable using large hardware infrastructure, modern software engineering tools for big data processing. The contribution of this work can be characterized as follows:

- We go beyond a simple binary classification by introducing ArabFake, an aspect-based fake news detection model that is able to coarsely tell not only if the given set of text content includes any star-category-like rumors but also in predict pairs (e.g. Fabricated-Story/Partisan) and associated risk level.
- We utilize the MARBERTv2 model, which is pre-trained on a massive collection of multi-dialectal Arabic tweets. This allows ArabFake to properly consider Arabic's complexity, considering its dialect variants, complex morphology, and informal nature popular on social media.
- ArabFake uses a multitask learning framework, allowing the model to jointly learn and perform three tightly connected tasks (i.e., detection, categorization, and risk assessment). This enables learned knowledge to be more easily shared across multiple tasks, yielding better model performance and efficiency.
- We introduce a new resource named ArabFake Dataset which is hand-labeled by experts moreover the sparsely optically recognized data in the domain. It contains high quality labels for fake news categories and levels of risk, which makes this dataset a valuable resource that will support future work related to Arabic Fake News Detection.
- We include valence scores in ArabFake so that we can reveal the most characteristic words and expressions of fake news, which makes our model more transparent with some linguistic patterns and manipulative tricks employed in Arabic fakes news.

## II. RELATED WORK

Detecting misinformation is challenging due to complex content, requiring manual intervention for accurate automated methods. The methodologies and key findings from recent studies on misinformation detection in the Arabic language are summarized in Table 1. Researchers employ AI and deep learning to detect misinformation online [18]. Mubarak, et al. [19] proposed a novel approach using emojis to identify offensive and hate speech in Arabic tweets. They collected 12,698 Arabic tweets with offensive emojis, manually annotated them for different offensive levels, and achieved competitive performance on external datasets. The study provides linguistic analysis and insights into cultural and contextual factors influencing hate speech detection in Arabic. By utilizing the extralinguistic data in emojis, Mubarak et al. [19] suggested a novel method for identifying offensive and hateful tweets written in Arabic. The performance of the suggested technique is competitive on external datasets, indicating that the gathered information captures traits common to all foul language. Additionally, the authors offer linguistic analyses and insights into the contextual and cultural elements that affect the recognition of abusive language and hate speech in Arabic.

**TABLE 1.** This table summarizes key methodologies and findings in recent misinformation detection studies in the Arabic language.

| Reference | Methodology | Dataset/Corpus | Key Findings |
|---|---|---|---|
| Rangel et al. (2019) | LDSE method for deceptive tweets | Diverse Arabic corpora, false COVID-19 information dataset | LDSE method for deceptive tweets, dataset support for Arab researchers |
| Ozbay and Alatas (2020) | Two-step method for fake news detection | Not specified | Document preprocessing, performance evaluation of 23 algorithms |
| Madani et al. (2021) | Algorithm for fake news identification in Morocco | Not specified | NLP, ML, and DL-based algorithm, acceptable misinformation detection accuracy |
| Ameur and Aliane (2021) | Dataset creation for misinformation detection | 10,828 fake Arabic tweets | Dataset creation process, potential use for hate speech |
| Haouari et al. (2021) | LDSE on diverse Arabic language corpora | Diverse Arabic corpora, false COVID-19 information dataset | LDSE method for deceptive tweets |
| Mubarak et al. (2022) | Emojis for offensive/hate speech detection | 12,698 Arabic tweets with offensive emojis | Competitive performance, cultural and contextual insights |
| Himdi et al. (2022) | Supervised ML for fake news detection in Arabic | Arabic fake news articles dataset | Dataset creation, features extraction (emotion, linguistics, etc.) |
| AlKhamissi and Diab (2022) | Hate speech detection in Arabic tweets | Not specified | MTL models, AraHS ensemble, self-consistency checking, outperformed reference models |
| Jarrahi & Safari (2023) | AI and deep learning for misinformation detection | Not specified | Challenging nature, need for manual intervention |
| Sofy et al. (2023) | Data mining, ML, classification for job ad fraud | EMSCAD dataset, Arabic version | RF classifier accuracy of 97%, importance of certain features |
| Abd Elaziz et al. (2023) | Multitask learning, meta-heuristic for COVID-19 news | Arabic social media data using a pre-trained transformer | Outperformed other algorithms in accuracy, precision, recall |
| Mughaid, et al. (2023) | NLP, ML | online news sources, Twitter, Facebook, Instagram, WhatsApp, and YouTube comments | CNN outperforms SVM, Naïve Base, Logistic Regression, Random Forest, Gradient Boosting |
| Hashmi et al. (2024) | FastText embeddings with ML/DL models, including transformers. Applied regularization, hyperparameter tuning, and explainable AI techniques (LIME, LDA). | WELFake (72,134 news articles) FakeNewsNet (23,196 records) FakeNewsPrediction (6,335 instances) | CNN-LSTM with supervised FastText embeddings achieved best performance (0.99 accuracy). Improved upon baselines, outperformed transformers, and provided interpretability through LIME and LDA. |
| Mallik & Kumar (2024) | Integrated Word2Vec embedding with stacked LSTM and dense layers for context-free fake news detection. | Kaggle Fake News (20,800), TI-CNN Synthetic (20,015), ISOT Fake News (44,898), SMS Spam (5,572). | Proposed model achieved 98-99% accuracy across datasets, outperforming baselines and generalizing well to different domains. |

Sofy et al. [20] created an intelligent Arabic model utilizing data mining, machine learning, and classification methods to identify fake job advertisements. Different classifiers, including SVM, RF, NB, and KNN, were utilized to apply their proposed model to the EMSCAD dataset to identify fraudulent jobs. The analysis revealed that the RF classifier had the best accuracy, reaching 97%. In the study's experiments with various features and algorithms, the processing text of benefits feature considerably increased the model's accuracy. The study found that several characteristics, including "company logo," "company profile," "interview questions," "benefits," and "department," are important in identifying employment fraud. Another study conducted by Elaziz et al. [21] proposed a methodology that uses multitask learning and meta-heuristic algorithms to identify fake Arabic news related to COVID-19. Their system extracts contextual features from Arabic social media using a pre-trained transformer-based model and a modified Fire Hawk Optimizer for feature selection. The methodology outperformed other algorithms' accuracy, precision, recall, and F-measure, achieving a 59% accuracy rate in identifying COVID-19-related misinformation.

Rangel et al. [22] and Haouari et al. [23] employed the Low Dimensionality Statistical Embedding (LDSE) method on diverse Arabic language corpora to identify deceptive tweets. Elhadad et al. [24] curated a dataset of false COVID-19 information to support Arab researchers.

Subsequent research developed a machine-learning model for detecting false COVID-19 information, utilizing different machine-learning algorithms using several feature extraction methods [24]. Ozbay and Alatas [25] used a two-step method to present a fake news detection model. The initial step involved data preprocessing to remove redundant terms and characters, constructing a document terms matrix, and assigning weights to the terms in each document. The second step involved evaluating the performance of twenty-three artificial intelligence algorithms and conducting a comparative analysis. Himdi et al. [16] focused on detecting fake news in Arabic using supervised machine-learning model that categorizes Arabic news items based on the reliability of their context. The authors introduced an Arabic fake news articles dataset created via crowdsourcing. They developed an Arabic Natural Language Processing tool to extract four textual features: emotion, linguistics, part of speech, and polarity. The model achieved an accuracy of over 75% in predicting the veracity of Arabic news articles, with linguistic features proving to be the most influential cue in detecting misleading text.

Ameur and Aliane [26] formulated a misinformation dataset containing 10828 fake Arabic tweets to detect misinformation in the Arabic language. Another data set collected is the Arabic Fake News Dataset (AFND). With the aid of supervised and unsupervised machine learning techniques, this dataset intends to assist academics in classifying the veracity of Arabic news items. The data set includes 606912 openly available news articles from 134 news websites in 19 Arab nations. Using an Arabic fact-checking program, the news source was categorized as reliable, not credible, or uncertain.

Wotaifi and Dhannoon [27] applied deep learning techniques in order to detect Arabic fake news using the AraNews dataset. The dataset includes Arabic articles from various fields like politics and culture. The researchers proposed a Hybrid Deep Neural Network that combines the Text-Convolutional Neural Network (Text-CNN) and Long Short-Term Memory (LSTM) architectures. The Text-CNN identifies key features, while the LSTM handles long-term dependencies. The results of the study indicated that the hybrid model achieved an accuracy of 0.914, outperforming the Text-CNN and LSTM, which had accuracies of 0.859 and 0.878, respectively surpassing the performance of previous studies using the same dataset.

Madani et al. [9] developed an algorithm to identify fake news on social media in Morocco. A categorization method based on natural language processing, machine learning, deep learning, and novel tweet properties made up the algorithm. The adopted model showed acceptable results, as the misinformation detection accuracy was equal to 79%. Mubarak et al. [19] introduced a method for collecting many offensive and hate tweets in any language using the extralinguistic information contained in emojis. The method was applied to Arabic tweets, and the resulting dataset was annotated and publicly released. Then, machine learning models for identifying offensiveness and hate speech were trained on the dataset and evaluated. External datasets from Twitter, YouTube, and Facebook were also used to test the models' generalizability. By utilizing the data gathered by this method, the researchers discovered and captured the universal traits of hostile language and identified frequent terms, targets, and patterns in offensive communications. It also highlighted the limitations of natural language processing models, such as their inability to consider cultural context or understand nuances like sarcasm. Moreover, our model address other limitations in the previous models such as the limited scope, Arabic language challenges, and limited explainability.

AlKhamissi and Diab [28], developed NLP models for hate speech detection in Arabic tweets for three separate tasks: determining whether a tweet contains offensive language, recognizing hate speech, and classifying the type of hate speech. Using MTL training, they created a model known as AraHS, an ensemble of MARBERT models optimized for the fine-grained hate speech subtask. According to the study, the model performs better than the reference models. The models outperformed earlier techniques by 3.4%, achieving an accuracy rate of 82.7% for the hate speech subtask. The study also suggested a self-consistency correcting technique that raises performance even higher. In conjunction with the preceding studies outlined in the related work section, the investigation conducted by Mughaid et al. [29] contributes to this evolving landscape by introducing a dual-component framework employing NLP and ML techniques. This framework extends the scope of data collection from diverse sources, including online news and various social media platforms, and integrates state-of-the-art ML algorithms for cyberbullying detection. Notably, their evaluation results highlight the superiority of convolutional neural networks (CNN) in this context.

Misinformation detection in Arabic is a large landscape with multiple innovative techniques being used by researchers and different datasets covered. ArabFake, in this paper, is a contribution to the synthetic category space growth by stemming from more traditional supervised learning on fake news detection and performing it using multiple approach deep learning framework which was sufficed for identifying complexity of detecting and categorizing then risk assessing whole categories. There have been previous studies that look at these parts individually, but ArabFake takes a new approach using Multi-Task Learning (MTL) to improve across many other aspects. In the following, we present background with a description of our architecture and datasets for ArabFake; while at the same time presenting evaluation results to demonstrate its success in solving multi-faceted challenges faced by misinformation in Arabic content.

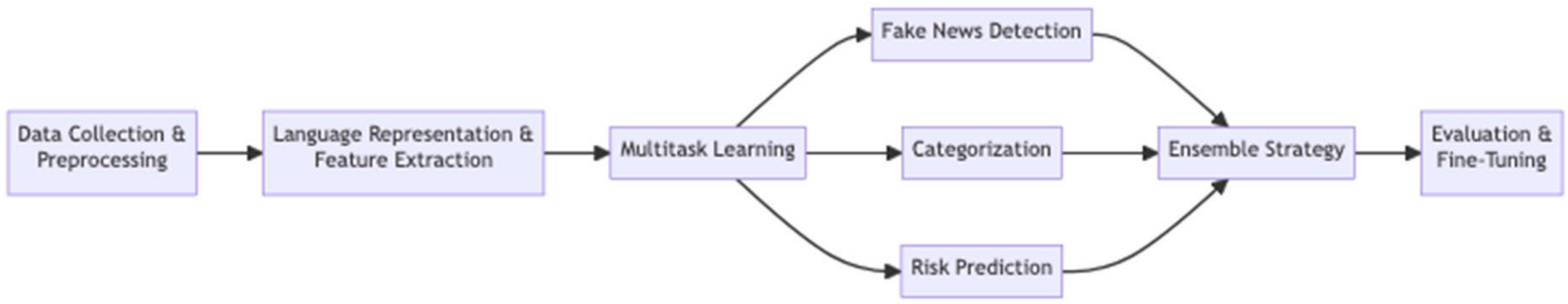

**FIGURE 1.** ArabFake architecture consists of three main tasks, fake news detection, categorization, and risk prediction.

## III. ArabFake: ARABIC FAKE NEWS DETECTION FRAMEWORK

This section describes the ArabFake methodology that we follow for detecting and mitigating the proliferation of fake news within the Arabic language context. The data collection, data pre-processing, and word embedding techniques we employ to train the suggested model, as well as other learning models for comparison, are covered in this section.

### A. MOTIVATION AND OBJECTIVES

Fake news presents a substantial risk to reliable information and societal trust. Disseminating false information can significantly influence public opinion, undermine democratic processes, and decrease public confidence in media sources. Consequently, there is a need to develop robust and accurate fake news detection methodologies to counteract the detrimental impact of misinformation. Addressing this issue poses inherent challenges, owing to the dynamic and sophisticated tactics employed by fake news makers, the large volume of data requiring analysis, and the complicated linguistic nuances, particularly when dealing with languages like Arabic. This paper presents the development of ArabFake, ArabFake is a novel Arabic Fake News Detection framework that utilizes state-of-the-art language models for detecting fake news in Arabic text. ArabFake addresses several limitations of current state-of-the-art models for Arabic fake news detection. Unlike many models that focus solely on binary classification, ArabFake offers a more comprehensive approach by also categorizing the type of fake news (e.g., political, health-related) and predicting associated risk levels as shown in Figure 1. It leverages the MARBERTv2 model, which is better suited to handle the linguistic diversity and informal nature of Arabic social media content. Additionally, ArabFake improves explainability by using valence scores to highlight the words and phrases indicative of fake news, providing insights into linguistic patterns. Its multitask learning framework allows for joint training on detection, categorization, and risk assessment, enhancing performance and efficiency. Finally, ArabFake introduces a new manually annotated dataset, the ArabFake Dataset, to address the scarcity of labeled data, contributing a valuable resource for future research. The integration of multitask learning facilitates the concurrent execution of interrelated tasks, culminating in heightened overall efficiency and accuracy in detecting fake news in the context of the Arabic language.

### B. DATA COLLECTION

A new dataset, "ArabFake Dataset," was formulated by extracting a subset of the AraNews dataset [30], which contains fake labels. The subset selection was based on the characteristics with a keen focus on articles that contain potentially fake or misleading attributes. This selection of the dataset was based on specific criteria such as diversity of topics, sources of information, and the representation of Arabic news. After selecting the subset of fake news, a category label is assigned to each article for annotation. Labels were created for each statement within the dataset before selecting the statements. However, after conducting thorough research and reviewing the already available literature, it can be said that it is important to label each statement based on its category before executing further analysis. These labels help to identify the classification of the statements as per their nature, such as political, economic, fabricated, or satire [31], [32], [33]. These labels were chosen because of the description they give out for the subject matter of fake news. We have assigned the most suitable category label for each fake news article with respect to its content.

Similarly, we assigned a risk level label to fake news to categorize the risk level annotation based on how misleading news impacts the public. As per research, there are significant impacts of fake news, such as anxiety or mental stress, impact on decision making, and facing difficulties while finding authentic or trustworthy news [34], [35]. We also concluded that fake news, other than causing mental stress, can also cause harm to people in other ways. It has also been found that fake news contributes to a trend that reduces trust in official informational sources.

**TABLE 2. Samples of fake news tagged subclasses.**

| Label: Category | Label: Risk | Example |
|---|---|---|
| Fabricated stories | Risk | القنصلية العامة الامريكية بالدار البيضاء , اكدت , من جانبها , في اتصال مع " المساء " , انها الغت فعلا بعض تاشيرات المغاربة التي تخول لهم السفر نحو الولايات المتحدة الامريكية , بعد ان ظهر انها غير مستوفية لجميع الاجراءات المتعلقة بإصدارها .<br>"The US Consulate General in Casablanca confirmed, in contact with 'Al Massa' newspaper, that it has indeed canceled some visas for Moroccans that allow them to travel to the United States, after it was found that they did not meet all the requirements related to their issuance." |
| Economy | No-Risk | واكدت رئيسة كوريا الجنوبية بارك جيون هاي علي الشراكة الاقتصادية الاقليمية الشاملة خلال اجتماعها مع رجال اعمال يابانيين يوم الاربعاء في البيت الازرق الانتخابي .<br>South Korean President Park Geun-hye emphasized comprehensive regional economic partnership during her meeting with Japanese businessmen on Wednesday at the Blue House, the presidential residence. |
| Political | Risk | 2 % وفقا لوزارة الشؤون الاقتصادية الروسية , مع انكماش الناتج المحلي الاجمالي .وقد تراجع الدخل الحقيقي للمواطن الروسي العام الماضي بنسبة 8 الروسي بنسبة 8.8%<br>The real income of the Russian citizens declined by 8.2% last year, according to the Russian Ministry of Economic Affairs, alongside a contraction of the Russian Gross Domestic Product (GDP) by 8.8%. |

To annotate the dataset, an evidence-based strategy was drawn with respect to previous research conducted in the same literature to consider the possible effects of fake news on some people. All these efforts ensured that each statement in the news, articles, etc., is labeled according to their characteristics or risk level. Table 2 represents some examples of categories of fake news as per our customized annotation [36].

### 1) ANNOTATION PROCESS

To ensure consistency and reliability within the annotation process, we recruited three expert annotators to annotate over 2459 samples. This annotation task was done by assigning two labels to each fake news sample. One label was given to categorize the fake news, and the other was given for its associated risks. We also provided clear guidelines for the category and risk level annotation to understand and perform a comprehensive evaluation of the annotated data. IAA was also incorporated in this annotation process, measured using the Fleiss Kappa score. Combined data from all three annotators was collected for over 80% of the total fake news, and we achieved 72% of IAA for the entire dataset. The results were shown using the Fleiss Kappa score, which is based on a high level of agreement among annotators, hence validating the annotation process. A voting approach was adopted to assign labels to the fake news sample.

### 2) FAKE NEWS ANNOTATED DATASET

The distribution of fake news categories in our dataset, as illustrated in Figure 2 and Table 3, reveals significant patterns in the types of misinformation circulating in Arabic media. Fabricated stories form the largest category with 1,508 instances (60.4% of the dataset), indicating that completely manufactured content is the predominant form of fake news. This is followed by economics-related fake news with 558 instances (22.4%), suggesting that financial misinformation is the second most common type of fake content. Political fake news constitutes 340 instances (13.6%), reflecting the significant presence of political disinformation in Arabic media. Notably, health-related fake news (61 instances, 2.4%) and religious fake news (28 instances, 1.1%) represent smaller but potentially impactful categories. While these categories have fewer instances, their potential impact on public health decisions and religious communities makes them particularly concerning. This skewed distribution towards fabricated stories and economic content suggests that fact-checking efforts in Arabic media should particularly focus on these domains, while maintaining vigilance across all categories due to their varying potential for social impact. Figure 3. Distribution of risk levels in fake news content. The risk assessment distribution shown in Figure 3 and Table 4 reveals a concerning pattern in the potential impact of fake news in Arabic media. The dataset shows a clear predominance of high-risk content, with 1,632 instances (65.4%) classified as "Risk" compared to 863 instances (34.6%) classified as "No-Risk". This approximately 2:1 ratio of high-risk to low-risk content indicates that not only is fake news prevalent, but it tends to manifest in forms that pose significant potential harm to society. This risk distribution has important implications for content moderation and fact-checking priorities. The prevalence of high-risk content suggests that automated detection systems should not only identify fake news but also prioritize the detection and rapid response to high-risk content. The distribution across the training (1,218 risk vs 733 no-risk), development (192 risk vs 79 no-risk), and test sets (222 risk vs 51 no-risk) maintains similar proportions, ensuring that our model learns to identify both risk levels effectively while reflecting real-world distributions.

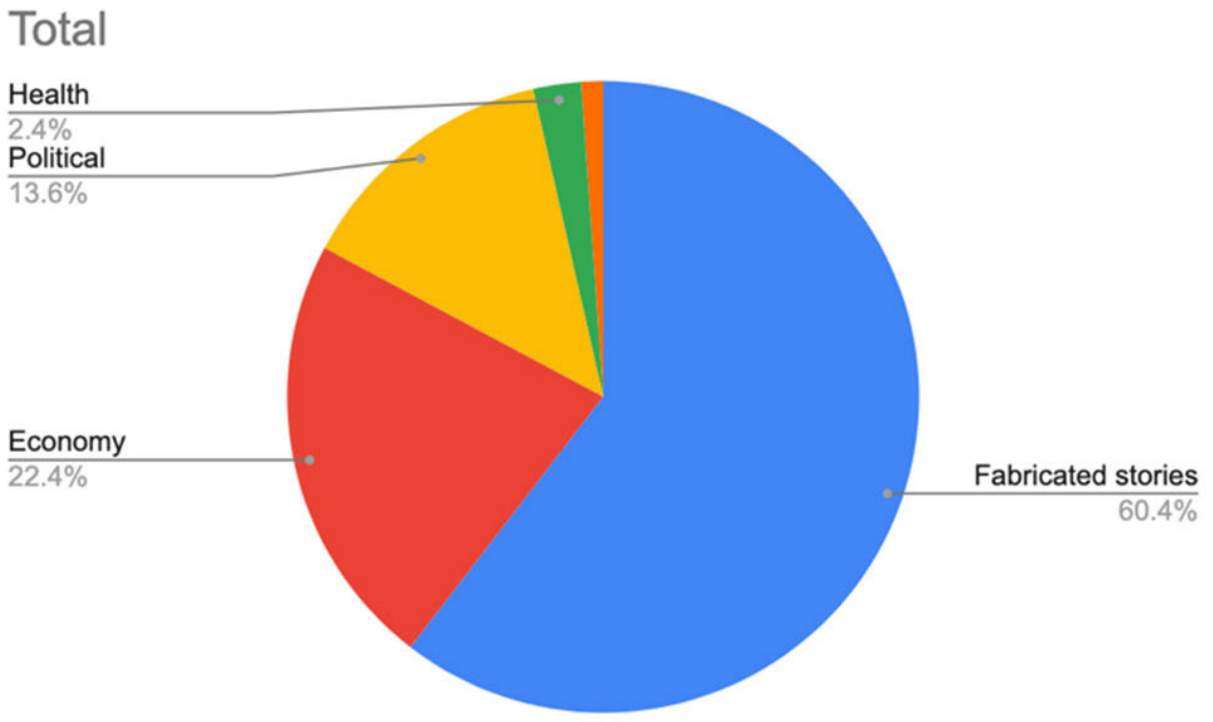

**FIGURE 2. Fake news categories.**

**TABLE 3.** Samples of fake news tagged subclasses.

| SUBCLASSES | TRAIN | DEV | TEST | TOTAL |
|---|---|---|---|---|
| FABRICATED STORIES | 1199 | 140 | 169 | 1508 |
| ECONOMY | 411 | 75 | 72 | 558 |
| POLITICAL | 272 | 47 | 21 | 340 |
| HEALTH | 46 | 5 | 10 | 61 |
| RELIGIOUS | 23 | 4 | 1 | 28 |
| TOTAL | 1951 | 271 | 273 | 2495 |

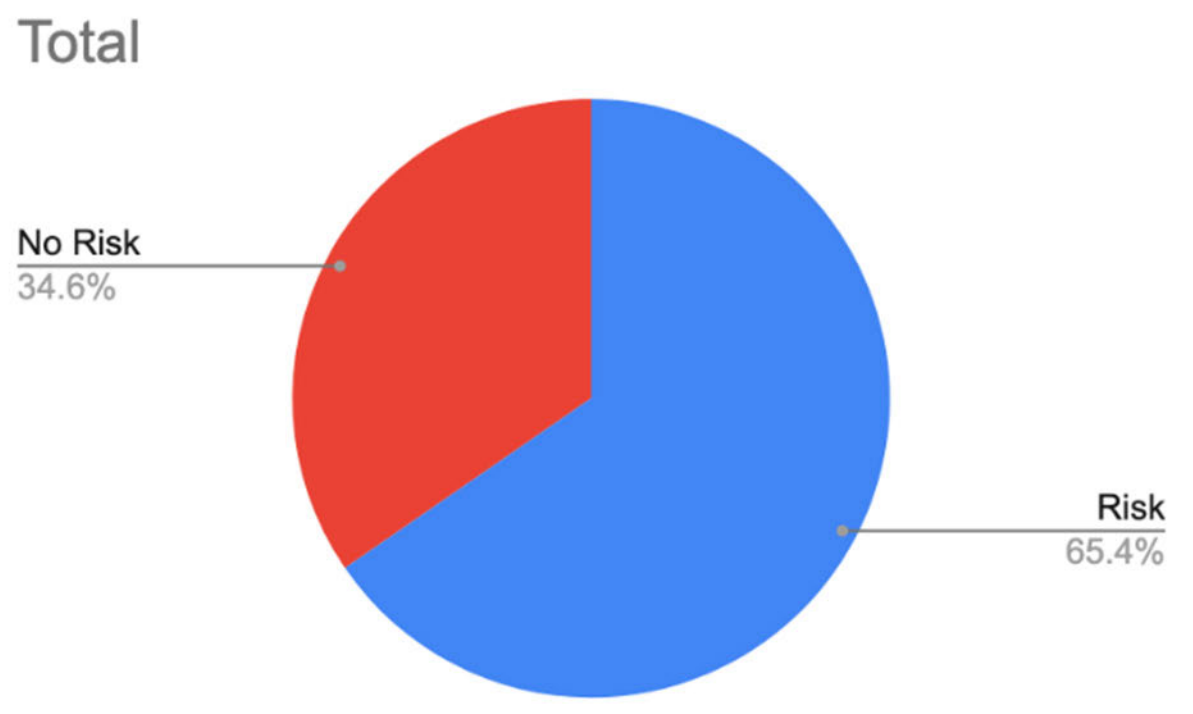

**FIGURE 3.** Fake news risks (Risk and No risks).

**TABLE 4.** Fake news risks.

| SUBCLASSES | TRAIN | DEV | TEST | TOTAL |
|---|---|---|---|---|
| RISK | 1218 | 192 | 222 | 1632 |
| NO RISK | 733 | 79 | 51 | 863 |
| TOTAL | 1951 | 271 | 273 | 2495 |

### C. DATA PREPROCESSING

Data preprocessing is an essential part of ArabFake's methodology. The preprocessing began with text normalization, where diacritics, Tatweel (i.e., elongation characters), and non-standard Arabic letters were removed or standardized to reduce the data variability [37]. Following normalization, a robust cleaning process was applied to remove unnecessary characters, punctuation, and numbers that could introduce noise into the model. Additionally, common Arabic stopwords were filtered out to focus the model on more meaningful content. The confluence of these preprocessing techniques enhances ArabFake's proficiency in uncovering subtle linguistic nuances embedded within fake news articles, laying a robust foundation for subsequent analysis and interpretation phases [38].

### D. LANGUAGE REPRESENTATION AND FEATURE EXTRACTION

The ArabFake model is based on two major components, which are feature extraction and language representation. Both components form a well-developed approach to detecting and categorizing fake or misinformation. Language representation involves translating the information from textual form to a structured numeric format. This structured numeric format provides efficiency in computational analysis. To understand the linguistic representation, converting text into structured format plays a critical role in detecting fake news. One of the main components in the language representation is the word embedding. It is basically a set of dense packets of vectors [39], [40]. This form of dense vectors clarifies the meanings of words. The understanding of words results in creating embedded words that are close to each other. Previous research provides deep insights into the set of Arabic word embedding where SVM, CNN & LSTM models are specifically applied to them [41], [42]. On the other hand, some models are being trained on billions of words taken from Arabic Wikipedia, Arabic Tweets, and Arabic Treebank [43], [44].

Moreover, the pre-trained BERT models used tokenization to create embedded words [45]. For example, some models use sub-word tokenization, such as AraBERT, which is suitable for Arabic and Moroccan Arabic Fields [46]. On the other hand, models such as QARiB mostly use character-level tokenization for handling informal and colloquial Arabic text [47]. Some NLP applications, including sentiment analysis, text classification, and named entity recognition, can be used to fine-tune the word embeddings generated by the above-mentioned models. ArabFake model utilizes the power of advanced language models along with word embeddings to achieve a robust language representation. MARBERT, which is a contextualized embedding model, has been employed to encode words and phrases to make them dense vector Fields [48]. These vectors are then represented using sophisticated semantic and syntactic attributes. This embedding gives power to the ArabFake for detecting subtle patterns, coherence, and other linguistic cues hidden in the Arabic news articles for fake information.

The feature extraction process involved the identification as well as extraction of the attributes of vectorized representations. Some traditional methods, such as TF-IDF and n-grams, and other advanced techniques, such as contextual embedding, contribute to creating a feature matrix that encapsulates relevant linguistic characteristics [49]. These features represent the base of the classification of tasks, which further empower ArabFake to analyze and recognize the authenticity of news articles. After the recognition of such news articles, the model categorizes them and estimates the risks associated with them.

### E. MULTITASK LEARNING (MTL)

Multitask Learning (MTL) is a widely-used machine learning paradigm where the model performs concurrently multiple tasks [21], [48]. ArabFake uses the power of MTL to perform fake news detection, classification and estimation in Arabic language simultaneously. MTL allows the model to leverage these relationships and learn from one task while performing advantageously in other tasks. ArabFake manages a more accurate and functionally-superior version of the overall experience by knowledge, attribute and representation sharing between tasks for an even better handling in fake news detection complex scenario. This not only optimises for each individual task, but also models in the small interdependencies among them to ultimately give a more robust architecture that is effective at tackling fake news on Arab media.

The key characteristic of MTL, which is task-relatedness, plays a crucial role in ArabFake's effectiveness. These tasks are related because they all involve processing and understanding Arabic text data to discern fake from genuine news. By jointly learning from these related tasks, ArabFake can capture the shared patterns, linguistic nuances, and contextual information essential for accurate fake news detection, categorization, and risk assessment in Arabic media, as shown in Figure 4. The tasks addressed by ArabFake exhibit various degrees of relatedness, and this task-relatedness is a crucial factor in the framework's design. Here's how the tasks in ArabFake are related:

- Fake News Detection and Categorization: Fake news detection and news categorization tasks are closely related because both involve analyzing the content of news articles. The features extracted for identifying fake news can also be informative for categorization and vice versa. For example, certain linguistic patterns commonly found in fake news might also indicate the article's category. Additionally, categorization can provide context for assessing the authenticity of news—fake news articles may belong to certain categories more frequently.
- Fake News Detection and Risk Prediction: Fake news detection and risk prediction are interrelated tasks. ArabFake may consider factors contributing to the news's risk level when identifying fake news. For instance, news articles containing false information about health or public safety issues might be flagged as high-risk. Conversely, a high-risk prediction might trigger a closer examination of potential fake news.
- News Categorization and Risk Prediction: News categorization and risk prediction share a relationship, particularly in assessing the potential harm associated with news articles. Certain categories, such as "Politics" or "Health," might inherently carry a higher risk because misinformation in these domains can have severe consequences. ArabFake leverages this relationship to enhance its risk prediction by considering the news category as one of the risk factors.

### F. MODEL TRAINING AND FINE-TUNING

In this work, we fine-tuned MARBERTv2 as the core component of our models, a bidirectional contextual transformer that leverages the power of bidirectionality. It generates a word representation using the previous and next context. It was pre-trained on 1B multi-dialectal Arabic tweets. It used the "BERT-base network architecture" [46], which is 12-layer, 768-hidden, 12-head with 160 M parameters. The three subtasks representing the core of ArabFake architecture are illustrated in the winch and consist of the following three learning models. The first is a fake news detection model, which receives a statement as input and determines whether it is false or not through a binary classification task. In order to leverage advanced language models and feature extraction techniques, ArabFake utilise a binary classifier to assess the authenticity of a given news article. This task involves analyzing the text's linguistic patterns, contextual cues, and inherent biases to determine its veracity. ArabFake's second task involves categorizing detected fake news articles into five distinct categories: Fabricated stories, economy, politics, health, and religion. This categorization provides deeper insights into the nature of the fabricated content. The categorization component employs a multi-class classifier to label fake news. This task enhances the granularity of analysis, facilitating a nuanced understanding of the diverse strategies employed by fake news creators to manipulate information. The third task is concerned with quantifying the risk associated with each identified fake news article. It involves analyzing various factors related to fake news, such as the news source, content, and intended audience. This analysis is then used to determine the risk associated with fake news. We trained a binary classifier using a human-annotated risk assessment dataset that we built by expert annotators (i.e., see section). Sample output for a sentence using the three subtasks' models, Fake news, category, and risk models, is shown in Figures 5 and 6. This risk estimation empowers users to prioritize and respond to fake news based on its potential ramifications.

We were also careful to take into account the risks of dataset bias, and possible false positives or negatives during model training and fine-tuning. To alleviate these issues, we used a mix of techniques to make the model performance equal across classes. Some of these were oversample minority classes, using class weighing within the loss function and regularization to avoid overfitting. We were also able to conduct a significant amount of error analysis after training, in order correct any bias that was still remaining. These steps were critical in helping to ensure our model was resilient and dependable, constraining the extent of dataset biases influencing its predictive performance.

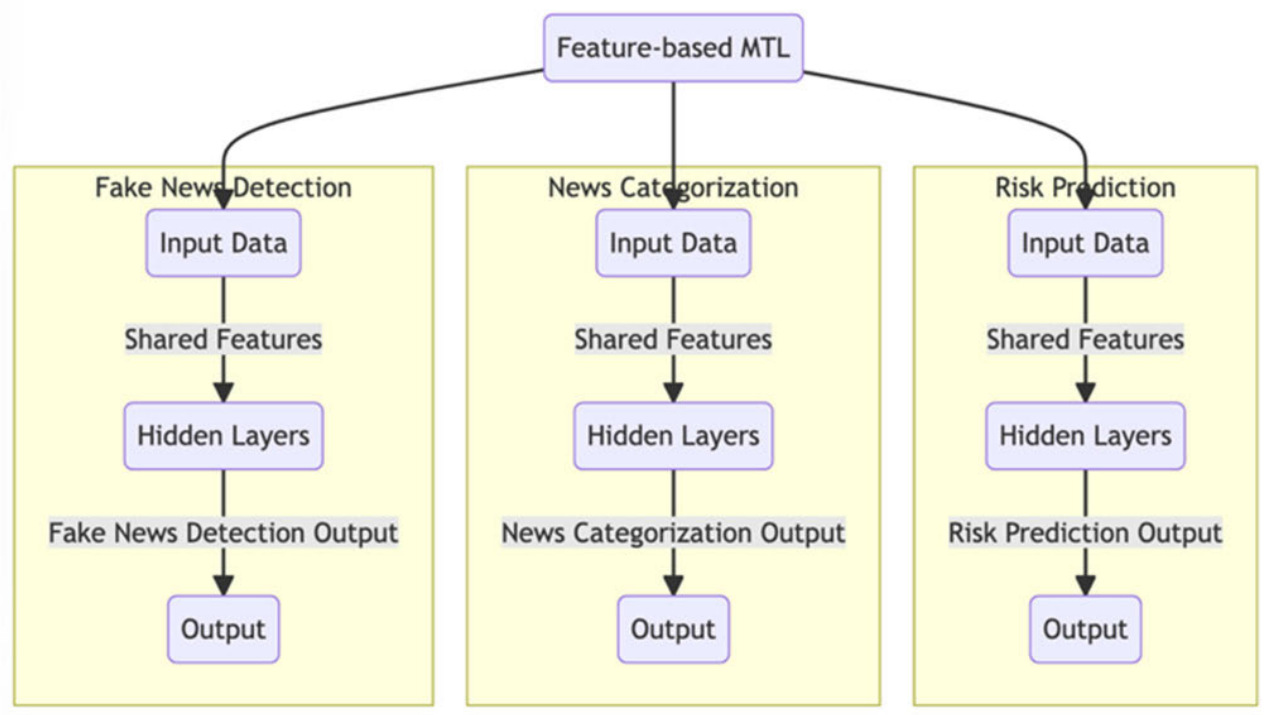

**FIGURE 4.** Shared features multitask learning for AraFake tasks: Fake news detection and categorization and risk prediction.

## IV. EXPERIMENTAL METHODOLOGY

### A. DATASETS

As our analysis ground, we have used two separate benchmark datasets with scrutinizing evaluation process upon which to build the effectiveness of ArabFake's Fake News Detection model: ANS corpus Khouja [51] and AraNews dataset [26]. It functioned as a large database against which we measured our Fake News Detection model. In contrast to how we evaluated the Category and Risk models, for ArabFake Dataset this time, we used our own crafted one. The data set made it possible for us test these particular components of our framework over a broad range. Spurring our analysis to be more robust and context-specific in nature, we therefore sought to enhance the ArabFake resource by creating a bespoke dataset from which an understanding of both categorization accuracy as well as risk prediction could take place. This two-way method in effort with benchmarks datasets and ArabFake Dataset yields a complete evaluation framework(Figure 5 and 6).

**Statement:** القنصلية العامة الامريكية بالدار البيضاء , اكدت , من جانبها انها الغت فعلا بعض تاشيرات المغاربة التي , " المساء " , في اتصال مع تخول لهم السفر نحو الولايات المتحدة الامريكية , بعد ان ظهر انها غير مستوفية لجميع الاجراءات المتعلقة باصدارها .
**English translation:** "The US Consulate General in Casablanca confirmed, in contact with 'Al Massa' newspaper, that it has indeed canceled some visas for Moroccans that allow them to travel to the United States, after it was found that they did not meet all the requirements related to their issuance."
**Classification: Fake statement**
**Category: Fabricated stories**
**Risk: It is a risk fake statement**

FIGURE 5. AraFake core three learning models.

#### 1) ANS CORPUS

Arabic News Stance is a dataset consisting of more than 4547 terms, and from them, 3072 terms are accurate while the remaining terms are fake [47]. Table 5 represents the stats for the dataset, which was created from samples taken from Arabic Text News Corpus, e.g., MENA region. The ANS corpus was created for the research community to have a tool to investigate fake information in the Arabic language. To create the ANS corpus dataset, a two-stage annotation process was involved where genuine statements from the Arabic language were extracted in the first step. In the second step, paraphrasing techniques were used to change the syntax of statements for creating fake data. This method proved helpful while ensuring that misleading statements have a high level of variation with respect to the linguistic representation of the dataset. Table 6 has a few examples of some fake news taken as a reference from Khouja [51].

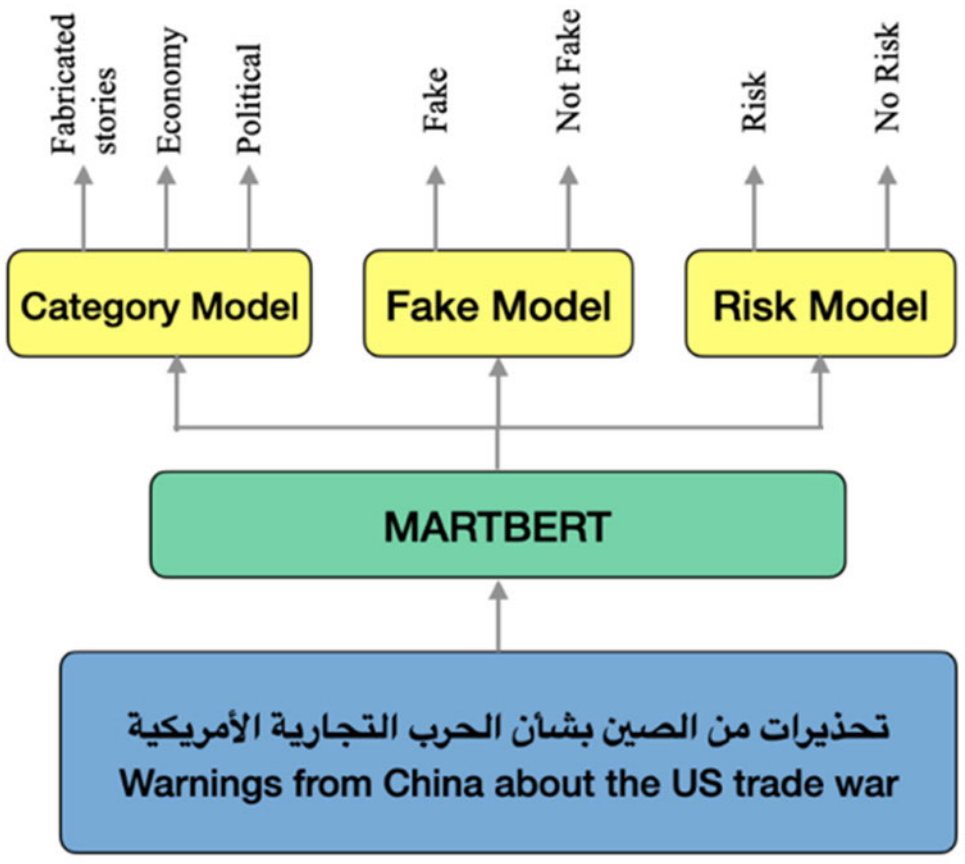

FIGURE 6. The sentence output using subtasks' models in AraFake.

#### 2) AraNews

The AraNews dataset provides researchers and interested people with an enormous database of Arabic news articles with many issues from several Arab countries [30]. It is supposed to be the producers' dataset to provide a useful resource to researchers, scholars, and experts working or even willing to work on natural language processing issues related to false information and fake news in Arabic. Due to the misleading information distributed in Arabic, millions of news stories comprise the dataset was planned to offer a wide-ranging and accurate sample of how the information got spread. An actual method should ensure the highest possible standard of the dataset's labels. Therefore, the authors carried out human annotation research. Thus, the study aims to reveal the task a group of annotators achieved by locating and labeling fabricated and modified news stories in the dataset. With the responsibility of detecting modified and false text by what the dataset makes of appropriating for developing and testing algorithms, this strategy was used. So, it is aimed that such a study provides training and assesses the constructed model's performance by utilizing the dataset's zero-shot version. The zero-shot version of the dataset is a portion of the complete dataset, including an equal number of true and misleading news stories, shown clearly in Table 5. It is resulted that the model in a regulated and representative setting can be trained and assessed.

TABLE 5. The statistics of used datasets.

| Dataset | Real | Fake | Total |
|---|---|---|---|
| ANS Corpus (Khouja, 2020) | 3072 | 1475 | 4547 |
| AraNews (Elmadany et al., 2020) | 97454 | 97330 | 194784 |

**TABLE 6.** Real and fake news adapted from Khouja [47].

| Label | Example |
|---|---|
| Real | تحذيرات من الصين بشأن الحرب التجارية الأمريكية<br>(Warnings from China about the US trade war) |
| Real | إيران تضع شروطاً لالتزامها بالاتفاق النووي<br>(Iran sets conditions for its commitment to the nuclear agreement) |
| Fake | تقارب وجهات النظر بين الولايات المتحدة وإيران<br>(The convergence of views between the United States and Iran) |
| Fake | قامت إسرائيل بالسماح بمرور الوقود إلى غزة<br>(Israel agreed to allow the transportation of fuel into Gaza) |

## B. EXPERIMENTAL SETUP

The original divides were used for both the ANS and AraNews datasets. The split for the ANS dataset was as follows: 70% for the training set, 20% for the development set, and 10% for the test set. A training set comprised 80% of the AraNews dataset, a development set of 10%, and a test set of 10%. The implementation uses deep learning tools like Keras and Tensorflow, together with the Python API. The AraBERT and QARiB models were used and modified to be adjusted for the same experiment with transformers to be compared with the recommended model ArabFake. Both models can be found on HuggingFace [52]. With a rate of learning of 2e-5, a batch size of 16, and a maximum sequence length of 128, it was possible to finetune AraBERT for five epochs. The original BERT model specification, which includes three epochs, a learning rate of 5e-5, and a batch size of 32, was used to fine-tune QARiB.

The ArabFake framework was tested for computational efficiency on a MacBook Pro M3 Max, where it demonstrated the ability to fine-tune models quickly using its 30-core GPU and hardware-accelerated ray tracing. The framework efficiently handled large datasets, thanks to the 16-core Neural Engine and high memory bandwidth. Performance metrics indicate that the framework can scale effectively with increased data sizes, maintaining high throughput without significant degradation in speed.

## C. EVALUATION METRICS

The effectiveness of fake news detection models can be evaluated by using a variety of evaluation metrics. Depending on the chosen metrics, the precise objectives and conditions of the task, as well as the features of the dataset, will all be used.

The following widely used evaluation metrics in this work have been taken into consideration, where each measurement is based on the macro-average [58].

- Accuracy: It calculates the ratio of properly categorized examples to the total number of instances. It is to distinguish correctly between fake and genuine news.
- Precision: measures the model's ability to avoid false positives, as it represents the proportion of correctly classified fake news instances out of all the instances predicted as fake news.
- Recall: assesses the model's ability to identify and capture as many instances of fake news as possible; it is the fraction of properly categorized false news incidents in the dataset out of all genuine fake news instances.
- F1 Score: It is the harmonic mean of precision and recall which of a false news detection model's overall effectiveness.

## D. BASELINE MODELS

### 1) mBERT

is a pre-trained multilingual model designed by Al-Twairesh [49]. It employs the Bidirectional Encoder Representations from Transformers (BERT) architecture, which is a neural network model for natural language processing. The mBERT model was trained on a large corpus of text in 104 languages. It consists of 12 layers, 768 hidden units, 12 heads, and 160 M parameters. The mBERT model can perform various linguistic tasks in various languages without requiring specialized training. Some of these tasks include text categorization, making the mBERT model a highly recommended tool for cross-lingual applications that require users to process text in many languages.

### 2) RoBERTa

is a Facebook AI Research (FAIR) pre-trained language model, which was trained on a corpus of about 160 GB derived from sources like books and web pages. Utilizing dynamic masking strategies during training greatly improved its competencies in diverse NLP processing tasks such as question answering and natural language inference. The architectural design includes either 12 or 24 transformer layers, resulting in an impressive total of approximately 1.5 billion parameters.

### 3) AraBERT

is a model based on the transformer model just for the Arabic language. This model involves BERT based transformer, which is actually a pre-training model for the Arabic corpus [46], [47]. The BERT transformer has 12-layer, 12-head construction and 768 hidden with 110M specifications. In order to enhance the pretraining process, a complete work masking methodology has been used specifically for this model. To do so, over 70 million sentences in the Arabic language were used to train the AraBERT model [53] along with the information available on Arabic Wikipedia, e.g., Open Source International Arabic New Corpus, as well as the Arabic Corpus [54].

### 4) QARiB

is also a model based on BERT, which has been developed specifically for the Arabic language [55]. This model has been designed to handle named entity recognition. Sentiment analysis and text recognition for a few of NLP tasks. This model has been trained using a substantial set of Arabic tweets. This model can be applied to several Arabic NLP applications, such as sentiment analysis text categorization. Topic modeling, etc. This model is also capable of handling informal and colloquial Arabic text because it was trained on a big corpus of Arabic tweets.

## V. RESULTS

### A. FAKE NEWS DETECTION RESULTS

Table 7 and Figure 7 present and compare the performance of other fake news detection models for two distinct datasets, e.g., ANS Corpus and AraNews. These models are specifically developed and trained to utilize advanced learning algorithms and natural language processing techniques. The model's success is based on key metrics such as accuracy, recall, F1 score, and precision. The developed model ArabFake is benchmarked against popular models, including mBERT, RoBERT, QARiB & AraBERT. The ArabFake model gave output with 73% accuracy on the ANS Corpus dataset, with 69% precision, 69% F1 score, and a 68% recall.

These results support the effectiveness of ArabFake model to detect fake news. We did that by comparing the performance of ArabFake to other models we have tested. ArabFake, however, performs the best in case of AraNews dataset (73%=>94%) indicating that no other model except ArabFake can identify fake news effectively in Arabic. ArabFake has produced strong performance in both datasets compared to other models, with similar results. This also shows that the model can detect fake news written in Arabic. We have demonstrated that this is the ideal model to comprehend fake news and its issues. It is difficult to keep the Arabic media clean and Arabcheck helps this happen.

These findings show that the effectiveness of fake news detection algorithms might vary greatly depending on the dataset being used. The results of the models using the AraNews dataset are higher overall metrics than the ANS dataset. The good results of using the AraNews dataset can be reflected in its size, which is much larger than the ANS dataset, as shown in Table 7. Using a larger dataset, such as AraNews, has greatly increased the performance of all models by around 24% on average. ArabFake performs comparably to the other models, mBERT, RoBERTa, AraBERT, and QARiB, on the ANS Corpus dataset but outperforms them on the AraNews dataset. This demonstrates the strength of ArabFake's multitask learning framework, which leverages the diversity and scale of the AraNews dataset to excel across tasks like fake news detection, categorization, and risk assessment. We found that AraBERT slightly surpasses ArabFake in terms of precision on binary classification. This improvement in precision can be attributed to AraBERT's single-task optimization, which effectively reduces false positives. On the other hand, AraFake, leveraging a multitask learning framework, achieves a balanced performance across multiple tasks, including detection, categorization, and risk assessment. While this multitask approach slightly affects precision in binary classification, it enhances overall versatility and performance across all tasks. This trade-off highlights the strength of AraFake as a comprehensive solution for Arabic fake news detection.

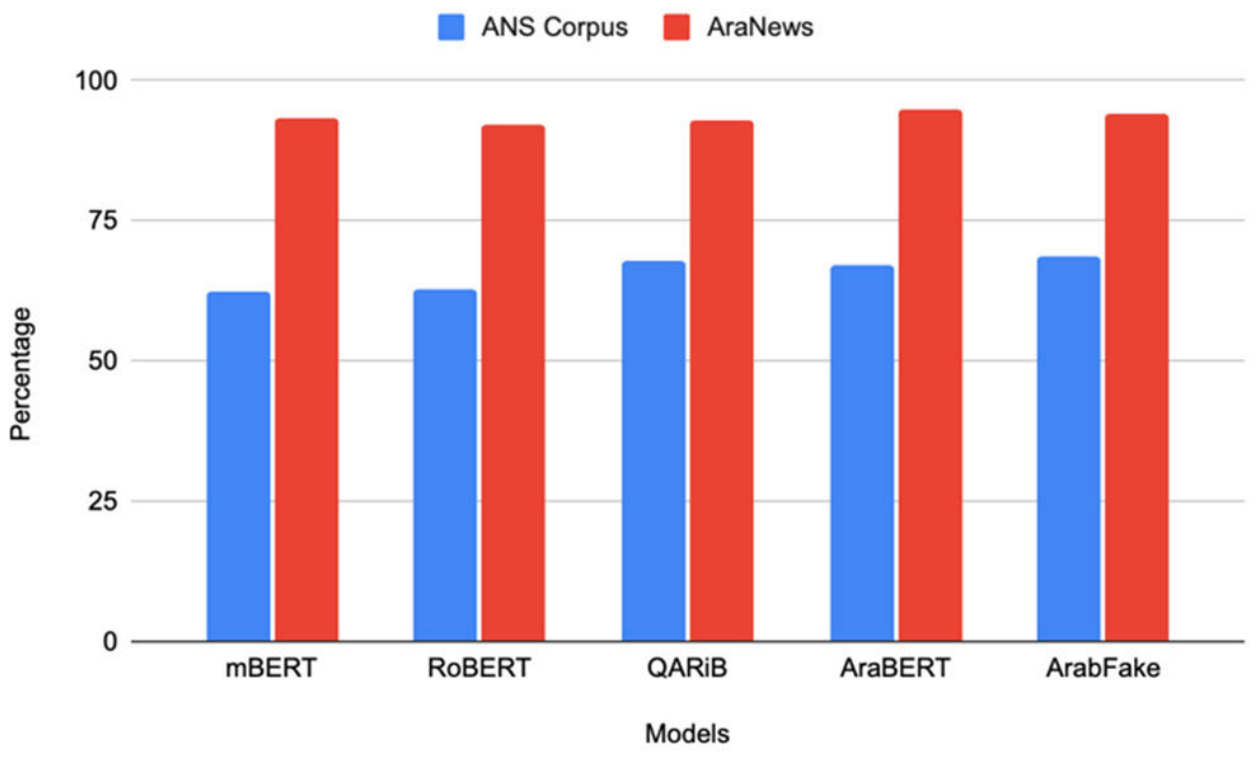

**FIGURE 7.** Fake news detection results of AraFake in compare with the baseline models.

**TABLE 7.** The results of the fake news detection models using two different datasets.

| Data | Models | Accuracy | Precision | Recall | F1 score |
|---|---|---|---|---|---|
| ANS Corpus | mBERT | 72.00 | 66.00 | 59.00 | 62.30 |
| | RoBERTa | 69.79 | 66.40 | 59.41 | 62.71 |
| | QARiB | 73.00 | 69.00 | 67.00 | 67.99 |
| | AraBERT | 73.00 | 69.00 | 65.00 | 66.94 |
| | **ArabFake** | **73.24** | **69.49** | **68.00** | **68.74** |
| AraNews | mBERT | 92.57 | 93.69 | 92.59 | 93.14 |
| | RoBERTa | 92.00 | 92.00 | 92.00 | 92.00 |
| | QARiB | 93.06 | 91.83 | 94.18 | 92.99 |
| | AraBERT | 94.58 | 97.13 | 92.39 | 94.70 |
| | **ArabFake** | **94.07** | **94.17** | **94.08** | **94.12** |

### B. FAKE NEWS CATEGORIES RESULTS

We fine-tuned mBERT, RoBERTa, QARiB, and AraBERT and compared them with ArabFake to assess the model's performance that classifies fake news into three categories: Fabricated stories, Economy, and Political. As shown in Table 8 and Figure 8, the models were evaluated on the

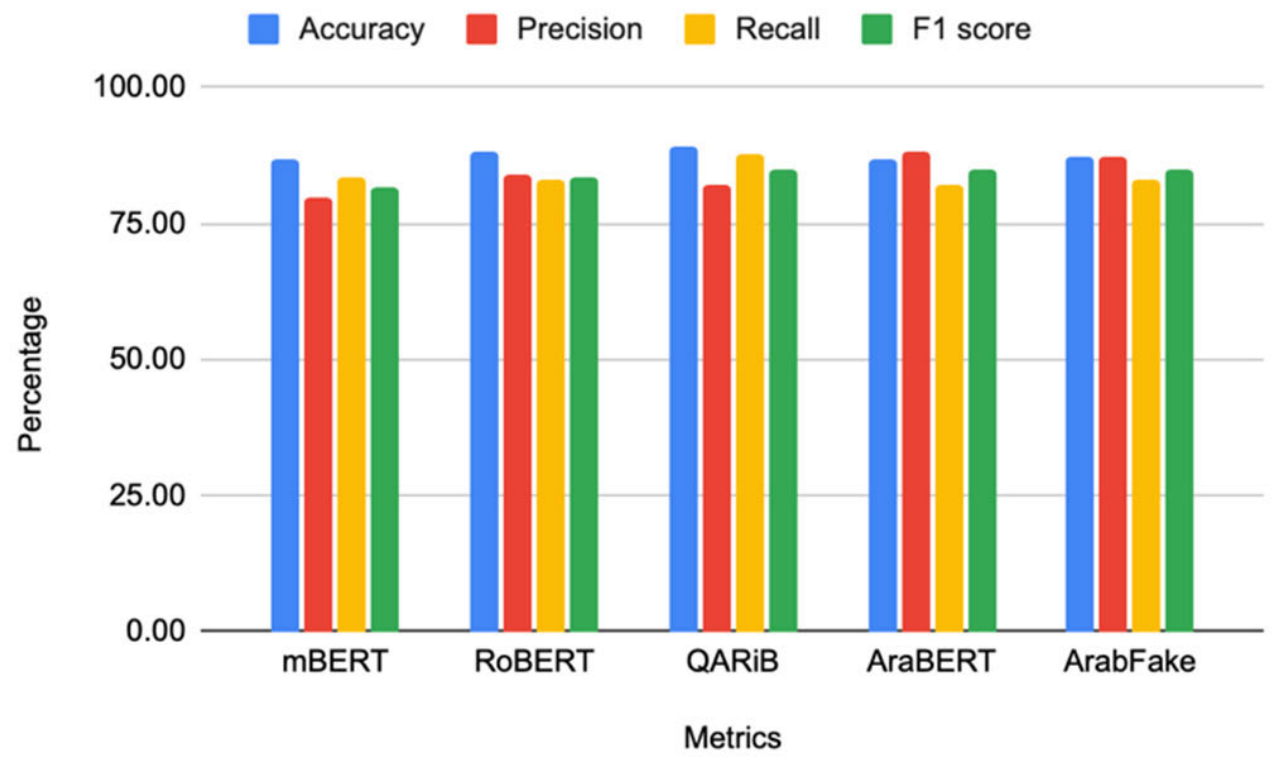

**FIGURE 8.** Fake news category results of AraFake in compare with the baseline models.

**TABLE 8.** The results of the fake news categories detection models using Annotated AraNews.

| Data | Models | Accuracy | Precision | Recall | F1 score |
|---|---|---|---|---|---|
| Annotated AraNews | mBERT | 86.87 | 79.80 | 83.44 | 81.58 |
| | RoBERTa | 88.00 | 84.00 | 83.00 | 83.50 |
| | QARiB | 88.93 | 82.08 | 87.71 | 84.80 |
| | AraBERT | 87.00 | 88.00 | 82.00 | 84.89 |
| | **ArabFake** | **87.02** | **87.05** | **82.89** | **84.92** |

**TABLE 9.** The results of the fake news risks detection using annotated AraNews.

| Data | Models | Accuracy | Precision | Recall | F1 score |
|---|---|---|---|---|---|
| Annotated AraNews | mBERT | 80.74 | 80.60 | 79.25 | 79.92 |
| | RoBERT | 81.00 | 80.00 | 79.00 | 79.50 |
| | QARiB | 82.41 | 73.37 | 81.63 | 77.28 |
| | AraBERT | 85.97 | 85.13 | 79.67 | 82.31 |
| | ArabFake | **89.96** | **90.49** | **87.39** | **88.91** |

Annotated AraNews dataset (i.e., section IV) using four metrics: Accuracy, Precision, Recall, and F1 Score.

Among the models, ArabFake achieved the highest accuracy of 89%, indicating its ability to classify fake news categories correctly. ArabFake also demonstrated superior precision with a score of 90%, highlighting its capability to identify relevant fake news categories accurately. The model exhibited a recall of 87%, reflecting its proficiency in identifying all relevant instances of fake news categories. Moreover, ArabFake achieved an F1 score of 88%, indicating a balanced performance in precision and recall.

Comparatively, the other models also exhibited impressive performance, mBERT, RoBERT, QARiB, and AraBERT all achieved high accuracy and precision, demonstrating their effectiveness in fake news category detection. These models consistently achieved strong recall and F1 scores, indicating their ability to accurately identify relevant categories of fake news. While these models performed well, ArabFake outperformed them in accuracy, precision, recall, and F1 score. Its superior precision and high F1 score demonstrate its ability to strike a balance between accurately identifying fake news categories and minimizing false positives. The results underscore the effectiveness of our developed model, ArabFake, in detecting and categorizing fake news accurately, positioning it as a promising solution for addressing the challenges of fake news in the Arabic language.

### C. FAKE NEWS RISKS RESULTS

We modified mBERT, RoBERT, QARiB, and AraBERT, and we compared them to ArabFake to assess how well the fake news risk model performed in categorizing fake news based on risk. Accuracy, Precision, Recall, and F1 Score were the same evaluation metrics we utilized.

As shown in Figure 9 and table 9, QARiB scored an accuracy of 82%, as well as 73% precision, 81% recall, and a 76% F1 score. ArabFake functioned somewhat better with an accuracy of 90%, precision of 90%, recall of 87%, and an F1 score of 88%. AraBERT achieved an F1 score of 82% and accuracy, precision, and recall of 86%, 85% and 80%. The results show that the accuracy, precision, recall, and F1 score of all models were comparable. The finest balance between precision and recall was found in ArabFake, which had the highest precision and F1 scores of 90% and 88%.

The high accuracy and precision scores (i.e., >=86% and >=90%) indicate that the models correctly classified most news articles as risk or no-risk. The high recall scores (>=87%) show that the models correctly identified most of the risk news in the dataset. The high F1 score (>=88%) is a balanced measure of precision and recall, indicating that the models achieved a good balance between correctly identifying risk articles and minimizing false positives.

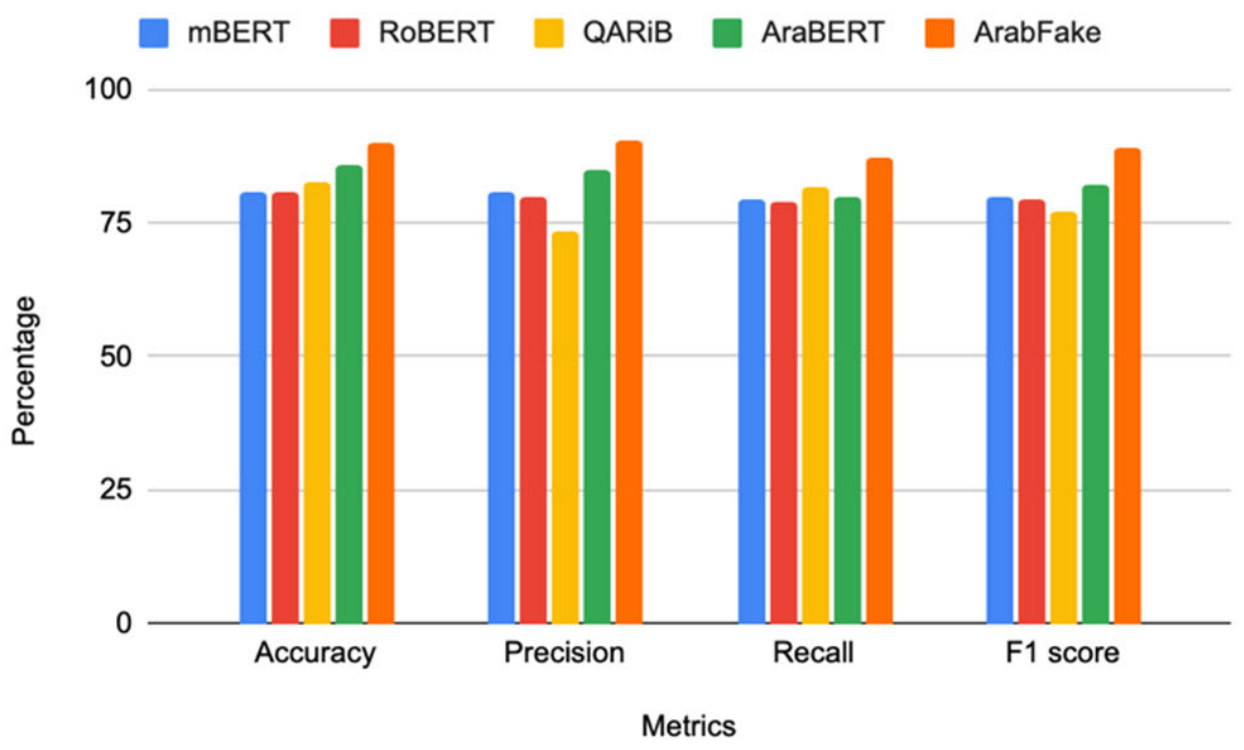

**FIGURE 9.** Fake news risk prediction results of AraFake in compare with the baseline models.

### D. FAKE WORDS VALENCE SCORE RESULTS

The valence score aids in determining a word's relative importance among false and true messages, according to Conover et al. [57]. We further estimated the valence score for words in fake and clean messages to study the substance of fake news. The valence score V(t) for a term t is calculated as:

$$V(t) = 2\frac{\frac{N(t,fake)}{N(fake)}}{\frac{N(t,fake)}{N(fake)} + \frac{N(t,clean)}{N(clean)}} - 1$$

where: - N(t,fake) = frequency of term t in fake news - N(t,clean) = frequency of term t in clean (legitimate) news - N(fake) = total number of terms in fake news - N(clean) = total number of terms in clean news The valence score has the following properties: - The score ranges from $-1$ to $+1$ - V(t) = $-1$ indicates term t appears only in clean messages - V(t) = $+1$ indicates term t appears only in fake messages - V(t) = 0 indicates balanced distribution between fake and clean messages For example, if a word appears 100 times in fake news out of 1000 total words (frequency = 0.1) and 20 times in clean news out of 1000 total words (frequency =

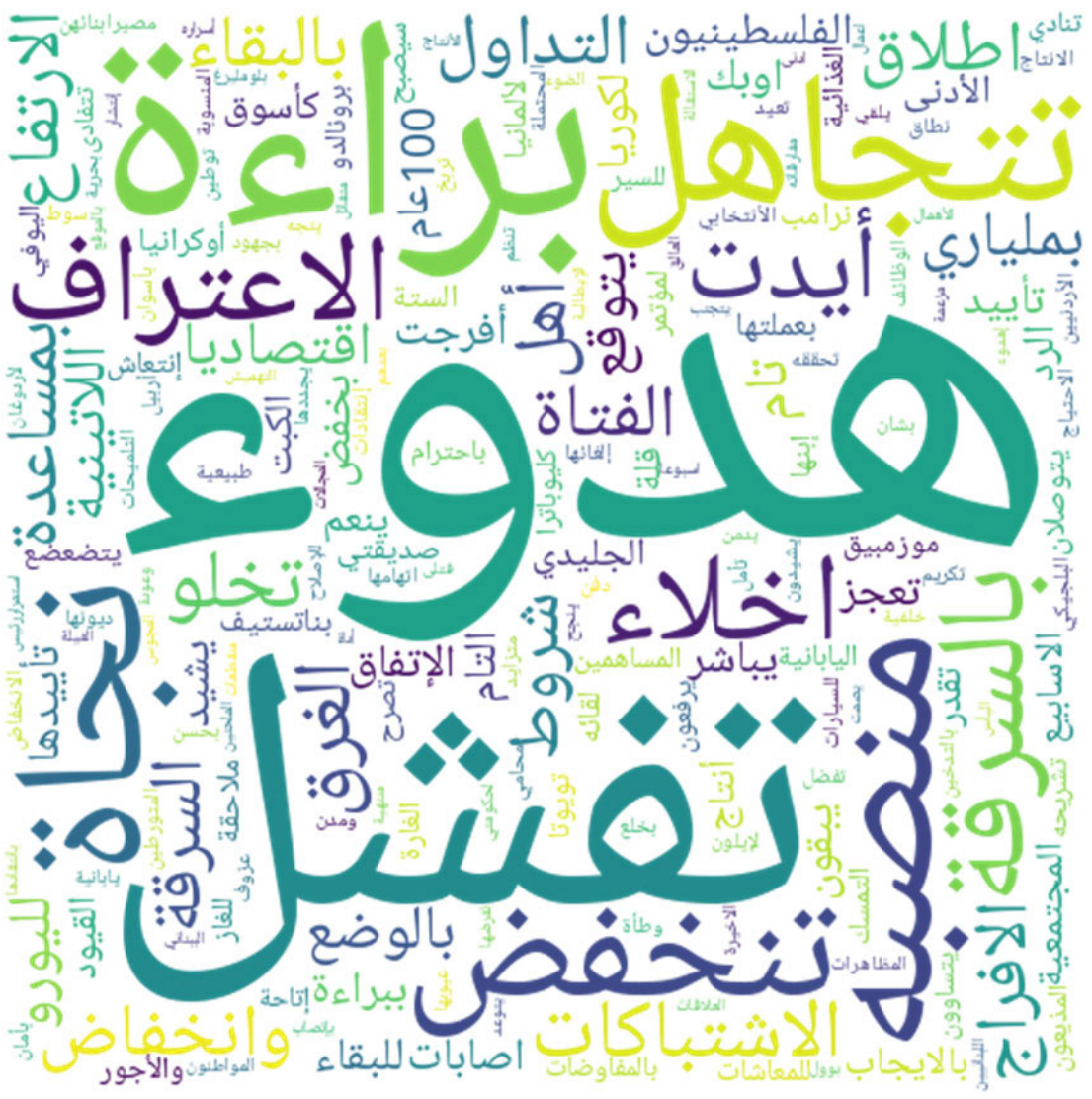

**FIGURE 10.** Tag cloud for words with top valence score among fake class.

0.02), its valence score would be approximately 0.67, indicating stronger association with fake news.

Figure 10 shows the word cloud for the highest valence score for words that appear in fake news. Where the larger fonts represent the words assigned with the highest valence scores, these words appear more in fake news than in the clean ones. If we look for the most prominent and significant words, we are going to find words such as هدوء (''Hodoa'' – ''Quiet''), تفشل (''tafshal'' – ''fail''), تتجاهل (''tatagahl'' – ''ignore''), براءة (''baraah'' – ''innocence''), تنخفض (''tankhfd'' – ''drop''), and اخلاء (''ikhlaa'' – ''evacuation''). Using such words can manipulate people's emotions or opinions. For instance, the word تفشل (''tafshal'' – ''fail'') was used to propagate this fake news: ''صاروخ استهدف نجران القوات السعودية تفشل فى اعتراض'' (i.e., Saudi forces fail to intercept a missile targeting Najran), this is classified as political high-risk fake news because it could have significant consequences that could damage the reputation and credibility of the Saudi military if readers believe the information to be true. This could lead to a loss of public trust and support for the government and military. It also could create fear and panic among residents of Najran, as they may believe that they are at risk of being targeted by missiles. This could cause social unrest and instability in the region. Spreading fake news can have serious consequences, especially when it involves sensitive topics such as national security. It is important to verify the accuracy of any news or information before sharing it with others and to seek out reliable and credible sources of information.

Another example that we obtained from the dataset is [تعرضت فيه لمحاولة الاغتصاب, دون ان يتدخل احد منهم لانقاذها, غير ان مقاومتها دفعت المتهم الي مغادرة المكان, سارقا منها هاتفها المحمول ومحفظتها ,.] Which translates, ''She was subjected to an attempted rape, without any of them intervening to save her. However, her resistance prompted the accused to leave the place, stealing her mobile phone and wallet.'' ArabFake classified this piece of news as political, high-risk fake news. The system identified it as fake news based on the use of certain words that are commonly associated with misinformation, particularly the word [اغتصاب] (''Eghtisab'' - ''rape''), which had a high valence score in the analysis of words frequently appearing in fake news. A third example [: أكسيا كابيتال ماركتس 88 شركة شحن مدرجة في نيويورك خسرت أكثر من 8.88 مليار دولار من الإيرادات في الأشهر التسعة الأولى] Which translates to ''Axia Capital Markets: 88 NY-listed shipping companies lost over $8.88 billion in revenue in the first 9 months'' ArabFake classified this piece of news as political, high-risk fake news. The system identified it as fake news based on the use of certain words that are commonly associated with misinformation, particularly the word [خسرت] ('''Khaserat'' - ''Lost''), [إيرادات] (' Eiadat'' – ''revenue''), which had a high valence score in the analysis of words frequently appearing in fake news.

## VI. DISCUSSION

Identifying fake news in this digital era became crucial due to the spreading of misinformation. This spread of fake news has graved social consequences and impacting our daily lives. In order to tackle this issue, ArabFake, which is a multitasking learning framework, has been developed to check and verify the news published on online platforms. Our ArabFake is developed using advanced techniques and models, which provide an impressive 94% accuracy rate for identifying fake news. This rating is based on the macro F1 score calculation. We used two benchmark datasets during our experiments, ANS Corpus and AraNews. Both contain different types of articles in Arabic.

To ensure the accuracy and dependability of our dataset, we also formulated a study with human annotators. For which a new dataset, ''ArabFakeDatset,'' was created via manual annotation by a team of experts. These experts classified news items based on their respective levels of danger. Our model adopts this approach to learn from a set of reliable and correct data to provide a strong foundation of training for the tool and for the assessment of news. During the test of our model on the annotated datasets, it has shown accuracy while identifying fake news and classifying it on its respective level. To do so, we also performed a comparison of our model's performance with other popular tools available in the industry. This was done to highlight the competitive advantage our model has among other fake news identifiers. With the solid and accurate results generated by our model, we can address various issues posed by fake news.

This research has several theoretical and practical implications for the detection of fake news in Arabic. The model generates a high F1 score while detecting fake news and furthermore builds its subtasks that makes the procedure for detection of fake notifications as per category wise along with risk associated to it. This is the reason why our model can be used in a real world. Such analysis illuminates the association between linguistic patterns and fake news articles, as well as other indicators. This offers some deep understanding which can be utilized for jeopardizing future interventions of media literacy efforts. The research suggests wide practical implications for news organisations, policymakers and other regulatory bodies to improve their policies on fake news dissemination. ArabFake, powered by high-performance hardware platforms and software frameworks like PyTorch and TensorFlow support efficient integration of the attributes of its use cases with computational efficiency aspects able to scale if needed. ArabFake is therefore approachable for real-world applications with large-scale datasets to be processed.

As we advance towards developing fake news detection model, there are several opportunities and challenges that need to be considered. Both of these were good examples to show how capable this application is in detecting the fake news posted online. This is a way of creating protective barrier to help to avoid the misinformation. Otherwise, there are some ethical problems being developed for the ArabFake as well. An ethical side of the ArabFake may present with how it develops: relying on a biased dataset is likely to produce false positives or negatives. Simply, it means news published online can be classified wrongly.

Not only but also Fake news detection models can possibly collect information about users, like in this case involving compromises such as social media activities and browsing history which is a big privacy concern. As a result, it leads us to data-driven decisioning in our approach that leans closer against users' control. By complying with human rights principles and data protection rules developers are taking care to balance the risks associated both by developing detectors as well as presenting these tools (fighting fake news, on one side, while protecting data of users which may suffer from different kinds of dangers).

ArabFake has several potential real-world applications in media and journalism, including automated fact-checking for news organizations, real-time social media monitoring to curb misinformation, and integration into media literacy tools to help readers assess the credibility of Arabic news articles. Additionally, it can assist investigative journalists in identifying disinformation patterns and support researchers and analysts in tracking and analyzing trends in Arabic fake news. These applications can enhance the accuracy of published news, empower users, and contribute to efforts against disinformation in the Arabic media landscape.

## VII. CONCLUSION AND FUTURE WORK

In this paper, we introduce ArabFake: a new multitask deep learning architecture for fake news detection in Arabic media. In our work, we showcase a substantive technical elevation where a multitask learning model is designed and fine-tuned that has achieved the best-ever accuracy in fake news detection now by reaching 94% accurate results. Using this framework, the methodology extends MARBERTv2 and includes a strategy to handle detection, categorization along with risk at the same instant.

This work provides a major construct of the ArabFake Dataset, consisting from 2,495 news item humanely annotated. The dataset was obtained by a reliable annotation process resulting in a high inter-annotator agreement 72% Fleiss Kappa score. This meticulously produced dataset serves as an important resource for subsequent research in Arabic fake news detection and offers a solid baseline to assess the performance of similar systems.

Our empirical analysis revealed important patterns in Arabic fake news distribution, identifying that fabricated stories (60.4%) and economic misinformation (22.4%) are the predominant forms of fake news in Arabic media. Notably, approximately two-thirds of fake news content poses high risk to society, underlining the urgency of effective detection systems. The implementation of valence scoring proved particularly effective in identifying linguistic patterns associated with fake news, providing insights into the characteristics of misleading content.

There is a wide range of practical use-cases for ArabFake and it has great potential to be used by fact-checking organizations, news agencies as well content-moderation systems, media-literacy initiatives, as well researchers/analysts to study disinformation patterns. The framework reduces the strain from a constantly growing pool of misinformation by simultaneously evaluating if content is true or not and also estimating what risk level we are looking at, which key for prioritising intervention as well as disproving high-risk misinfo quickly.

Nevertheless, there are still several limitations and future works that could be exploited. This work should be extended by future research to incorporate new Arabic dialects, topics categories and real time detection features; multimodal approaches including text-image-video fake news detections; etc. There are still some directions for enhancing the model's explainability, looking at transfer learning methods to low-resource Arabic dialects and adapting mechanisms to resist fake news patterns that may evolve with time.

Its reach extends more than achievements in the technical realm. The effectiveness of the ArabFake experiment is evidence that AI-driven technical responses to misinformation are worthwhile and demand a higher level of ethical conduct in automated content analysis. Subsequent work will need to balance the accuracy of detection with making it more computationally efficient, keep user privacy in check, and account for the myriad ways in which racism can manifest personally but also have cultural or even linguistic dimensions. To be optimally effective at preserving the integrity of information in online Arabic content spaces, sustainable integration with established fact-checking systems and finely tuning it to react and adapt against new varieties of misinformation will likely prove key for ArabFake.

**AHMED MAHER KHAFAGA SHEHATA** received the Ph.D. degree from Aberystwyth University, Wales. He is currently an accomplished Associate Professor of information studies with a distinguished 11-years career, demonstrating expertise in the impact of ICT on scholarly communication. He excels in scientific and technical skills, analytical thinking, and problem-solving. He has an impressive academic background with degrees in library science from Egyptian universities. He is also with Sultan Qaboos University, Oman, contributing to academia through teaching, research, and service. His research interests include information behavior, scholarly publishing, and digital libraries. He has contributed significantly to the field with numerous publications and has been recognized with awards, such as Sharjah 2018 Award of Library Literature.

**MOHAMMED NASSER AL-SUQRI** received the Ph.D. degree from Emporia State University, Emporia, KS, USA. He is currently a Professor of information studies, specializing in information technology. With a career beginning as a Demonstrator and progressing through roles such as a Lecturer, an Assistant Professor, and the Assistant Dean of the Training and Community Services, Sultan Qaboos University, Oman, his journey reflects a dedication to academia and community service. His research interests include human information behavior, knowledge management, and LIS education. He has published extensively on these topics and has been recognized with awards for his contributions to academic excellence and student supervision.

**NOUR ELDIN MOHAMED ELSHAIEKH OSMAN** received the Ph.D. degree in knowledge base management and decision support system from Multimedia University, Malaysia. With a career beginning at The Future University, Sudan, in 1997. He is currently a Dedicated Assistant Professor of information studies. He has extensive experience in teaching both undergraduate and postgraduate levels across various institutions. He has published significant research in knowledge-based systems and their application in higher education and SMEs, demonstrating his commitment to advancing knowledge in his field. His expertise includes information and communication technology, knowledge management, and digitalization.

**FATEN HAMAD** received the B.Sc. and M.Sc. degrees in computer science from the University of Jordan, and the Ph.D. degree in information science from Aberystwyth University, U.K. She is currently an esteemed Associate Professor of information studies. Her career spans roles from a Teaching Assistant to an Associate Professor at prestigious institutions, including the University of Jordan and Sultan Qaboos University. Her research interests include information retrieval, health informatics, and big data analytics. She has contributed significantly to academic libraries' understanding in Jordan through her publications and active participation in international conferences. Her dedication to education and research is further demonstrated by her involvement in community services and committee memberships, enhancing academic, and professional practices within her field.

**YOUSUF NASSER ALHUSAINI** (Member, IEEE) received the Master of Science degree in computer science (information management and security) from the University of Bedfordshire, U.K., in 2015. He is currently pursuing the Ph.D. degree in computing in Malaysia. He is an Academic Staff with Arab Open University, Sultanate of Oman Branch.

**AHMED MAHFOUZ** received the B.Sc., M.Sc., and Ph.D. degrees from Minia University, Egypt, in 2006, 2011, and 2017, respectively, in a joint supervision program with UBC, Canada. He is currently an Assistant Professor with the Faculty of Computer Science, AOU, Oman. He is interested in building intelligent security systems using the cutting-edge of machine learning and computer vision. His research interests include computer and information security, intelligent systems, authentication, biometrics, mobile, and ubiquitous computing. He has been involved in many research projects to protect smartphone users against a series of risks.